\documentclass[conference,compsoc]{IEEEtran}
\usepackage{blindtext, graphicx}

\ifCLASSINFOpdf

\usepackage[tight,footnotesize]{subfigure}
\usepackage{textcomp}
\usepackage{float}
\usepackage{graphicx}
\usepackage{epstopdf}
\usepackage{color}
\usepackage{wrapfig,lipsum}
\usepackage{setspace}
\usepackage{graphics}
\usepackage{grffile}
\usepackage{lipsum,multicol}
\usepackage{tabularx}
\usepackage{array}
\usepackage{algorithmic}
\usepackage{algorithm}
\usepackage{lineno}
\usepackage{mathptmx}
\usepackage{latexsym}
\usepackage{moreverb}
\usepackage{amsmath}
\usepackage{calc}
\usepackage{amssymb}
\usepackage{sidecap}
\usepackage{verbatim}
\usepackage{longtable}
\usepackage{multirow}
\usepackage{color,soul}
\usepackage{multirow}
\usepackage{microtype}
\usepackage{enumitem}

\definecolor{darkred}{RGB}{200, 50, 50}

\usepackage[style=ieee,minbibnames=1,maxbibnames=2,doi=false,isbn=false,url=false,eprint=false,date=year,bibencoding=utf8]{biblatex}
\AtEveryBibitem{\clearfield{pages}}
\AtEveryBibitem{\clearfield{volume}}
\AtEveryBibitem{\clearfield{number}}
\AtEveryBibitem{\clearfield{note}}
\AtEveryBibitem{\clearlist{language}}

\IEEEoverridecommandlockouts
\newcolumntype{C}[1]{>{\centering\let\newline\\\arraybackslash\hspace{0pt}}m{#1}}
\newcommand{\cmmnt}[1]{\ignorespaces}
\newcommand\Tstrut{\rule{0pt}{2.6ex}}         
\newcommand\Bstrut{\rule[-0.9ex]{0pt}{0pt}}   
\begin{document}
	
	%
	\title{
	\vspace{-0.6cm}
	{\fontsize{18.6}{23}\selectfont HDTorch: Accelerating Hyperdimensional Computing with GP-GPUs for Design Space Exploration}
	\vspace{-0.8cm}
	}

	
	\author{\IEEEauthorblockN{William Andrew Simon, Una Pale, Tomas Teijeiro, David Atienza}
		\IEEEauthorblockA{Embedded Systems Laboratory (ESL), Swiss Federal Institute of Technology Lausanne (EPFL), Switzerland\\
			\{william.simon, una.pale, tomas.teijeiro, david.atienza\}@epfl.ch}
		
		\vspace{-0.8cm}
		}
	
	\maketitle

	\begin{abstract}

		HyperDimensional Computing (HDC) as a machine learning paradigm is highly interesting for applications involving continuous, semi-supervised learning for long-term monitoring. However, its accuracy is not yet on par with other Machine Learning (ML) approaches. Frameworks enabling fast design space exploration to find practical algorithms are necessary to make HD computing competitive with other ML techniques.
		To this end, we introduce HDTorch, an open-source, PyTorch-based HDC library with CUDA extensions for hypervector operations. We demonstrate HDTorch's utility by analyzing four HDC benchmark datasets in terms of accuracy, runtime, and memory consumption, utilizing both classical and online HD training methodologies. We demonstrate average (training)/inference speedups of (111x/68x)/87x for classical/online HD, respectively. Moreover, we analyze the effects of varying hyperparameters on runtime and accuracy. 
		Finally, we demonstrate how HDTorch enables exploration of HDC strategies applied to large, real-world datasets. We perform the first-ever HD training and inference analysis of the entirety of the CHB-MIT EEG epilepsy database. Results show that the typical approach of training on a subset of the data does not necessarily generalize to the entire dataset, an important factor when developing future HD models for medical wearable devices.

	\end{abstract}
	
	\begin{IEEEkeywords}
		Hyper-Dimensional Computing, Machine Learning, PyTorch, GPUs, CUDA
	\end{IEEEkeywords}
	
	\IEEEpeerreviewmaketitle

	\section{Introduction}
	\label{sec:intro}
	
	Recently, HyperDimensional Computing (HDC) has emerged as an alternative Machine Learning (ML) framework to more traditional models such as random forests or neural networks, where its novel data representation strategy enables various advantages from both hardware and software perspectives. 
	HD computing has been used in a broad spectrum of applications, such as robotics~\cite{mitrokhin_learning_2019}, recommendation systems~\cite{guo_hyperrec_2021}, language recognition~\cite{karunaratne_real-time_2021} and more. Due to the current trends of using Artificial Intelligence (AI) and ML for personalized medicine~\cite{yu_artificial_2018} and wearable devices for health monitoring~\cite{sujith_systematic_2022}, many HDC biomedical applications have been proposed, varying from emotion recognition~\cite{chang_hyperdimensional_2019} and electromyogram gesture recognition~\cite{rahimi_hyperdimensional_2016} to epileptic seizure detection via EEG signals~\cite{burrello_ensemble_2021, pale_multi-centroid_2022}. 
	
	The highly parallel nature of HDC algorithms lends motivation to the development of specific HDC hardware accelerators. While such accelerators will certainly be implemented in future products that rely on HD computing, they are expensive and limited in algorithmic flexibility, a necessity for research into the HDC design space. Therefore, open-source, flexible GPU-accelerated HDC frameworks are necessary to enable efficient HDC research.
	
	\begin{figure*}
		\centering
		\includegraphics[trim={1cm 1cm 1cm 1cm},clip,width=0.95\linewidth]{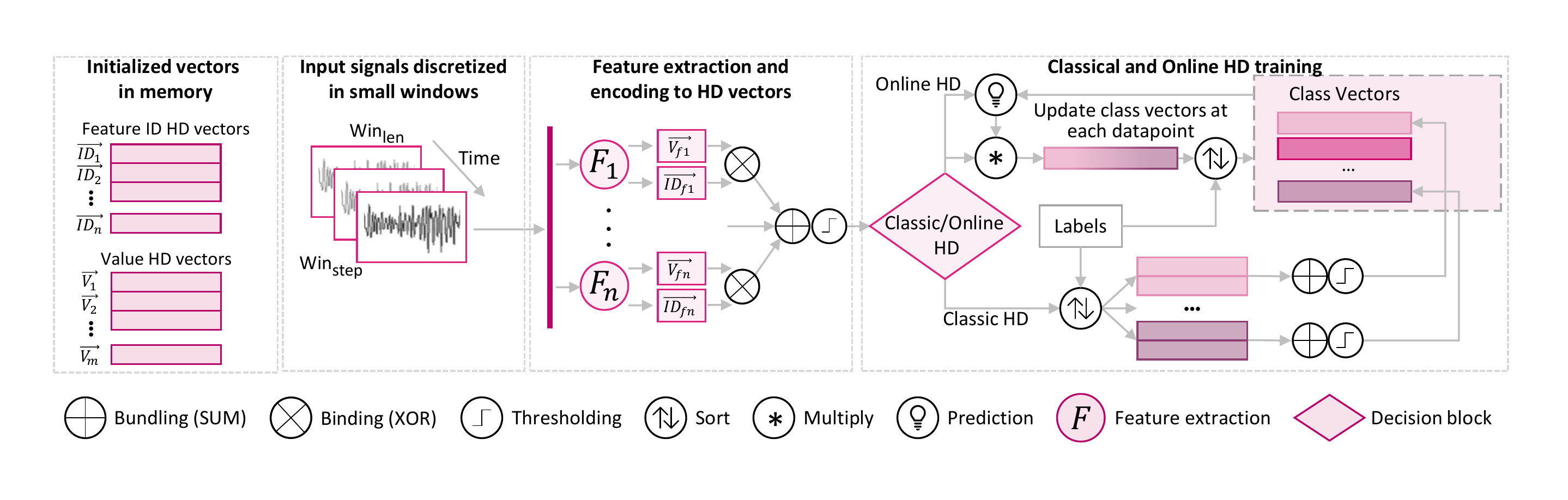}
		\caption{\small{HD workflow for training classical and online HD models. Online training differs in that the class vectors are updated after every datapoint by multiplying its similarity to the target class by the vector before accumulating it into the class.}} 
		\label{fig:hdworkflow}
	\end{figure*}
	
	In this context, we propose HDTorch, the first open-source, PyTorch-based library built for exploring the HDC paradigm. HDTorch unlocks the full potential of PyTorch applied to HDC algorithms, and further extends PyTorch with custom, CUDA-backed hypervector operations. HDTorch is highly customizable, enabling modification to hyperparameters and encoding/similarity strategies. We validate HDTorch's accuracy and runtime performance on four reference HDC benchmarks, demonstrating accuracy comparable to state-of-the-art works while greatly accelerating training and inference for classical and online HD strategies.
	
	We further motivate HDTorch's utility for exploring the HDC design space by applying it to the CHB-MIT epilepsy database. HDTorch enables us to perform the first ever HDC analysis of the entire dataset by reducing training and inference time by over 70x. We draw several conclusions from the analysis that will be useful for future works applying HD computing to large, unbalanced datasets.
	
	This work's contributions are summarized as follows:
	\begin{itemize}
		\item We introduce HDTorch, an open-source, PyTorch-based HDC framework with CUDA extensions for hypervector operations, namely, bit-(un)packing and bit-array summation in the horizontal/vertical dimensions.
		
		\item We benchmark classical/online HD computing with HDTorch, showing accelerations of (111x/68x)/87x for (classical/online) training and inference, respectively.
		
		\item We explore the accuracy/runtime impact of varying hyperparameters, namely, hypervector width and online HD training batch size. 
		
		\item We motivate HDTorch's design space exploration utility by performing, to the best of our knowledge, the first HDC training/inference analysis of the entire 980-hour, 7 million datapoint CHB-MIT epilepsy detection dataset. We present novel observations on the utility of HD computing on large, unbalanced datasets.
		
	\end{itemize}
	
	The remainder of the paper is organized as follows. Section~\ref{sec:related} details related work on HD computing. Section~\ref{sec:hdworkflow} provides a typical HDC workflow as motivation for HDTorch. Section~\ref{sec:HDTorch} presents the HDTorch framework. Section~\ref{sec:experimentalsetup} details the benchmarks, metrics, and testing environment we use to validate HDTorch. Section~\ref{sec:results} presents our experimental results, while Section~\ref{sec:conclusion} concludes this work.

	\section{Related Work} 
	\label{sec:related}

	\subsection{Hyperdimensional Computing}
	
	HD computing is a machine learning strategy whose defining feature is its representation of datapoints as long ('hyper') vectors, which enables learning by 'accumulation' of said vectors belonging to the same class. HD computing relies on two conditions; first, any two randomly generated HD vectors are with high probability orthogonal, and second, a vector generated by vector accumulation will be more similar to its components than a vector not of its class~\cite{kanerva2009hyperdimensional}. HD computing has proven to be very appropriate for various forms of learning such as online~\cite{moin_wearable_2021,benatti_online_2019} on-device, semi-supervised~\cite{imani_semihd_2019, pale_multi-centroid_2022}, and distributed learning~\cite{imani_framework_2019}. Storing models in the form of hypervectors exhibits strong noise and data corruption resistance~\cite{imani_exploring_2017}. It also enables exploration of feature importance and selection~\cite{pale_hyperdimensional_2022}.
	
	From an algorithmic perspective, many HDC variations have been explored in literature, touching on almost every aspect of the HD training or inference flow. These include methods for initializing hypervectors~\cite{imani_voicehd_2017,burrello_ensemble_2021, pale_systematic_2021}, accumulating datapoints into class vectors~\cite{imani_semihd_2019, pale_multi-centroid_2022, hernandez-cane_onlinehd_2021, kang_xcelhd_2022, rahimi_hyperdimensional_2020}, and calculating similarity between data and class vectors~\cite{kanerva2009hyperdimensional}. Exploring this design space necessitates analysis of the effects of various hyperparameter values such as hypervector lengths or online batch sizes, testing different encoding and learning strategies, and applying pre- or post-processing filtering to data or results. Unfortunately, due to the lack of publicly available libraries for fast processing and parallelization of HD computing on CPU or GPU, such analysis has thus far been performed on unoptimized HDC frameworks, greatly reducing research efficiency. For example, to the best of our knowledge, HD computing has not been tested on large datasets such as those for epilepsy detection. While previous works applied HD computing to epilepsy datasets, data subsets were always utilized~\cite{asgarinejad_detection_2020, burrello_laelaps_2019, burrello_ensemble_2021, pale_systematic_2021, pale_exploration_2022}. Unfortunately, it has also been demonstrated that training on subsets of data may result in significant alterations to the final predictions in comparison to utilizing the entire dataset~\cite{pale_multi-centroid_2022}. Thus, to develop suitable HDC algorithms, accelerating the exploration of the HDC design space is necessary.
	
	\subsection{Accelerating HD Computing}
	
	Several works have begun to explore strategies of accelerating HD computing via various software and hardware frameworks. Works mainly focus on 1) ASIC implementations~\cite{imani_revisiting_2021, imani_hierarchical_2018, wu_brain-inspired_2018, rahimi_robust_2016} or 2) in-memory computing accelerators~\cite{gupta_felix_2018, karunaratne_-memory_2020, karunaratne_energy_2021}. Other works explore GPU acceleration of HD computing. In~\cite{hernandez-cane_onlinehd_2021}, the authors implement an HDC architecture in PyTorch to compare against other ML algorithms on an embedded GPU, analyzing runtime, memory usage, and energy consumption. Similarly, the authors in~\cite{kang_xcelhd_2022} design a TensorFlow framework with HDC-specific extensions to accelerate the encoding and training process. Our proposal differs from the previous two in several ways. First, we utilize PyTorch as a base framework for HDTorch, as explained in Section~\ref{sec:HDTorch}. Second, as we want to encourage fast design space exploration in future works, we rely on native PyTorch operations where possible, and develop a set of custom CUDA functions where we identify HDC-specific bottlenecks that PyTorch cannot handle effectively. These functions are different from those accelerated in previous papers. Finally, we provide an open-source PyTorch library that encompasses these contributions.

	\section{Motivation: A Typical HD Workflow}
	\label{sec:hdworkflow}

	To motivate the need for HDTorch, we present a typical HD workflow that may be applied to a dataset by a researcher. The design choices presented here are commonly used in previous literature, and are accelerated by HDTorch, as described in Section~\ref{sec:HDTorch_extension}. 
	
	The first step when performing training or inference via HD computing is to encode raw input data into the HD space as hypervectors. One popular encoding method is that of ID-Level encoding~\cite{kang_xcelhd_2022}, where each feature has an ID vector representing it, while all datapoints are discretized into a fixed number of bins, with each bin having its own representative HD Value vector. ID vectors ($\vec{ID}$) are randomly generated, while Value vectors ($\vec{V}$) may be randomly generated or, as in~\cite{imani_voicehd_2017}, generated using a linear scaling method so that vectors representing similar values are also similar. Using the predefined $\vec{ID}$ and $\vec{V}$ vectors, each data point is encoded to a hypervector $\vec{H_{i}}$ by binding each feature vector $\vec{ID_{fi}}$ with the bin value vector $\vec{V_{fi}}$ corresponding to the value of the feature. Encoded features are then accumulated as formulated in Equation \ref{encoding}. The final summed values are normalized using majority voting to regenerate the binary vector. 
	\begin{equation}\label{encoding}
		\vec{H_{i}} = \Big \lfloor \sum_{f_{i}}{ \vec{ID_{fi}} \oplus \vec{V_{fi}} } \Big \rfloor
	\end{equation}
	
	Once encoded, the data can be processed in different ways. The most basic approach is so-called classical HD training, which utilizes single-pass accumulation of the encoded vectors belonging to the same class into a class vector. On the other hand, recent literature has proposed various improvements to classical HD training, such as iterative learning~\cite{imani_semihd_2019}, 'multi-centroid learning' which utilizes more than one vector per class~\cite{pale_multi-centroid_2022}, and progressive, online HD learning~\cite{hernandez-cane_onlinehd_2021}. In~\cite{pale_exploration_2022} the authors compared all these approaches on the use case of epileptic seizure detection and demonstrated that online training is potentially the most interesting, given its high performance, the fact that it utilizes each data point only once as opposed to iterative training, and its lower memory requirements than, for example, a multi-centroid approach. Online training improves accuracy by, instead of treating all data points as equally important, multiplying new data points by their similarity to current class vectors before being accumulated, as illustrated in Equations~\ref{eq:corrClassification} and~\ref{eq:wrongClassification}:
	
	\begin{equation}\label{eq:corrClassification}
		\vec{M_{C}'} \xleftarrow{}  \vec{M_{C}} + (\delta_{C}) \vec{H} 
		\vspace{-5mm}
	\end{equation}
	
	\begin{equation}\label{eq:wrongClassification}
		\vec{M_{W}'} \xleftarrow{}  \vec{M_{W}} - \gamma  (1- \delta_{W}) \vec{H} 
	\end{equation}
	where $\vec{M_{X}}$ is either the correctly or incorrectly classified class vector, $\delta_{X}$ is the distance from the class vector (lower distance means more similarity), and $\gamma$ is the learning rate. In this way, highly common class patterns are not allowed to saturate the class vectors, thus improving sensitivity to less common patterns. Online HD training is also applicable in continuously learning wearable devices, as it is capable of integrating new data in real time.
	The workflow is described visually in Figure~\ref{fig:hdworkflow}.
	
	Once class vectors have been learned from either the entirety of the training set in a single pass or over time via online training, test data can be compared to the class vectors to find the most similar vector, thus returning a predicted class. A variety of similarity metrics have been proposed, with the two most popular being Hamming distance and cosine similarity. 
	
	Then, let us consider a typical HD workflow that integrates the aforementioned steps. We will utilize as an example, subject 1 of the CHB-MIT epilepsy database consisting of 40 hours of data, split in 1 hour segments as in the original database, further described in Section~\ref{subsubsec:CHB-MIT}. We use the HDC framework from~\cite{burrello_laelaps_2019}, with ID-Level Encoding and Hamming distance similarity measurement. We perform both classical and online training using the common leave-one-out cross-validation strategy. 
	
	The results of training and inference are as follows. Classical training takes 46 minutes, 22 of which are consumed by the encoding step. Online training takes 134 minutes, 28 of which are consumed by encoding. If extrapolated to the full 980 hours of data, with a time resolution of 0.5 seconds, resulting in more than 7 million samples included in the CHB-MIT database, classical/online training would take 19/54 hours, respectively.
	
	It is no wonder, then, that previous works have analyzed only subsets of the dataset, as performing any sort of design space exploration is infeasible with such high runtimes. Thus, a GPU-accelerated, flexible HD computing framework is necessary to enable future fast, iterative research into the HDC design space.

	\begin{figure*}
		\centering
		\includegraphics[trim={1cm 1cm 1cm 1cm},clip,width=0.95\linewidth]{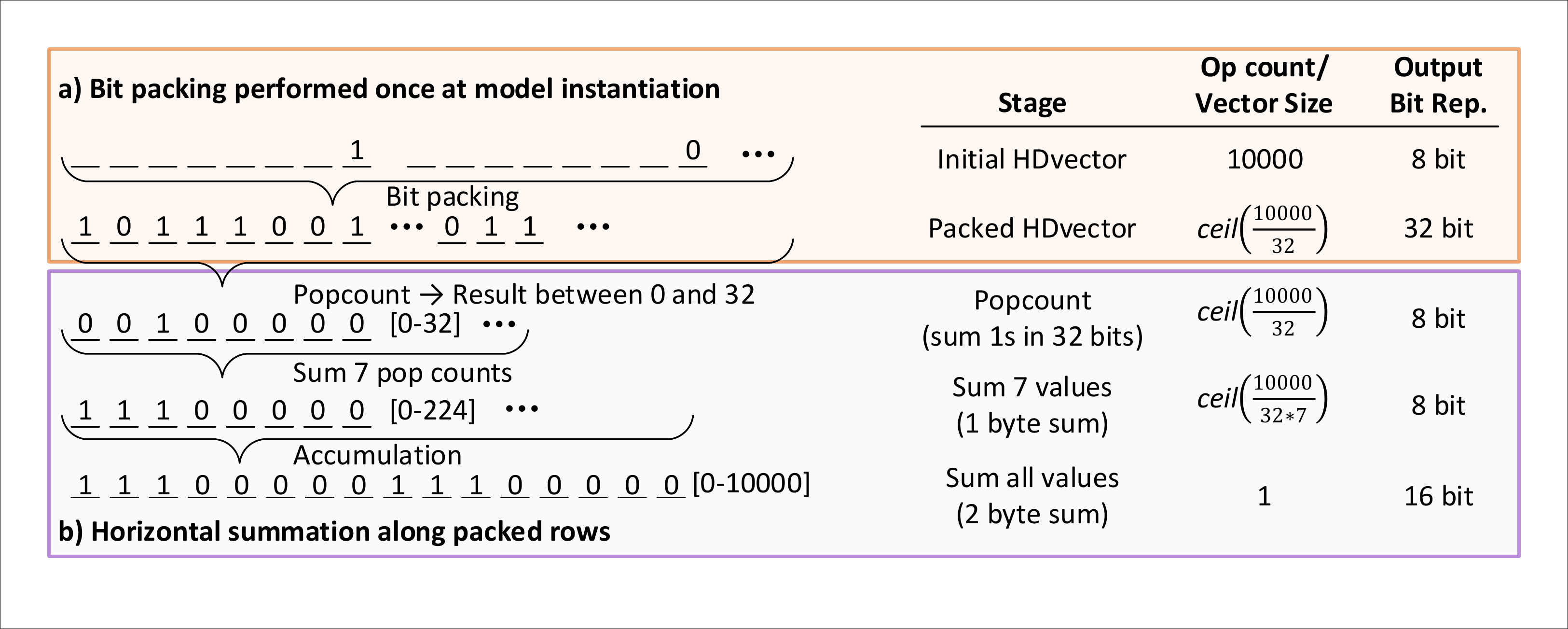}
		\caption{\small{Illustration of HDTorch's bit packing and horizontal summation operations. a) Bit packing reduces hypervector memory footprint and improves bitwise operation efficiency, and b) summation enables packed hypervectors to be accumulated efficiently.}} 
		\label{fig:hdextensions}
	\end{figure*}
	
	\section{HDTorch}
	\label{sec:HDTorch}
	
	
	With this motivation in mind, we describe how HDTorch provides a flexible framework for performing HDC research on GPU-equipped platforms.
	\begin{table}
		\caption{HDTorch Feature Overview} 
		
		\label{tab:HDTorchfeatures}
		\begin{tabularx}{\columnwidth}{ l l }
			\textbf{Features} & \textbf{Values}\\
			\hline
			Customizable & Hypervector Dimension \Tstrut\\
			Hyperparameters & Batch Size \Bstrut\\
			\hline
			Hypervector & Binary (0,1) \Tstrut\\
			Flavors & Bipolar (-1,1) \Bstrut\\
			\hline
			Hypervector & Random \Tstrut\\
			Generation & Scale Random~\cite{imani_voicehd_2017, pale_systematic_2021} \\
			Strategies & Sandwich~\cite{burrello_laelaps_2019} \Bstrut\\
			\hline
			\multirow{3}{*}{\shortstack[l]{Available\\Binding Strategies}} & ID-Level Encoding~\cite{kang_xcelhd_2022} \Tstrut\\
			& Feature Permutation~\cite{rahimi_hyperdimensional_2020} \\
			& Feature Appending~\cite{pale_hyperdimensional_2022} \Bstrut\\
			\hline
			Available & Hamming~\cite{kanerva2009hyperdimensional} \Tstrut\\
			Similarity Metrics & Cosine~\cite{kanerva2009hyperdimensional} \Bstrut\\
			\hline
			HD Computing & Binary (Un)Packing \Tstrut\\
			CUDA Extensions & Horizontal/Vertical Summation \Bstrut\\
		\end{tabularx}
	\end{table}
	
	\subsection{Implementing HD Computing in PyTorch}
	\label{sec:PyTorch}
	PyTorch is one of the two most popular Deep Learning (DL) frameworks utilized in research today, along with TensorFlow. \cmmnt{While similar in objective, the frameworks diverge along various lines, such as adoption in research and industry and model availability.} While the debate on framework superiority is contentious, there can be no doubt that PyTorch is currently the most popular DL framework in research, being utilized in over 75\% of research papers using either of the two frameworks~\cite{he2019mlframeworks}. As such, providing HDC support in PyTorch will enable the majority of the research community to explore HDC solutions on a wider range of topics more easily.
	
	HDTorch is realized as a python library installable from PyPI~\footnote{Omitted for blind review purposes.}. Table~\ref{tab:HDTorchfeatures} lists the features provided by HDTorch's base HD model class. Namely, it supports variable size binary or bipolar hyperdimensional vectors for any number of classes. Value and feature hypervectors may be generated randomly, or by a number of pseudo-random functionalities proposed previously in the literature~\cite{imani_voicehd_2017,burrello_ensemble_2021, pale_systematic_2021}. A variety of feature/class binding methods are supported, including ID-Level encoding~\cite{kang_xcelhd_2022}, vector permutation~\cite{rahimi_hyperdimensional_2020}, and feature appending~\cite{pale_hyperdimensional_2022}. Pairwise similarity calculation via either hamming distance or cosine similarity~\cite{kanerva2009hyperdimensional} is supported.

	\subsection{HDTorch: HyperDimensional Extensions for PyTorch}
	\label{sec:HDTorch_extension}
	
	While PyTorch's GPU optimization provides already excellent acceleration, its functions are not implemented with HD computing in mind and thus fail to fully optimize the HDC kernel. Namely, HD computing relies on operations between hypervectors containing 1000's of binary values. By default, PyTorch stores binary values as bytes, resulting in up to an 8x memory overhead per bit. Besides memory access, inefficiencies are also found in three key hypervector computations:  namely, 1) the performance of binary \texttt{xor} operations between hypervectors during both training and inference, 2) the summation of ID-Level encoded vectors for all features, and 3) the summation along the class hypervector during hamming distance calculation. 
	Encoding is usually the main bottleneck of HDC applications, taking up to 70\% of training time~\cite{kang_xcelhd_2022}, and is highly dependent on these inefficient operations. With this in mind, HDTorch provides four new functions with backing CUDA C++ code to address Python's shortcomings in regards to HD computing. These extensions are only applicable to binary hypervectors; bipolar hypervectors may still benefit from HDTorch's non-specific accelerations, as described in Section~\ref{sec:results}.
	
	\subsubsection{Bit (Un)Packing to Improve Memory and Bitwise Operation Efficiency}
	The first optimization HDTorch supports is the packing of a hypervector's bits into byte blocks. This reduces hypervector memory footprint by 8x and enables bitwise operations, specifically the bitwise \texttt{xor} necessary for ID-Value encoding and Hamming distance calculation. Individual bits are packed into 32-bit integers, enabling HDTorch to take advantage of CUDA's bit counting intrinsics as described below. A complementary unpacking operation is also supported to return packed hypervectors to their original state if necessary. Both operations are backed by CUDA code for GPU acceleration.
	
	\subsubsection{Horizontal/Vertical Summation for Highly SIMD Bitwise Operations}
	While PyTorch natively supports bitwise operations on packed bits, summation operations that can take advantage of the packed and binary nature of hypervectors are absent. Therefore, operations for performing horizontal and vertical summations are introduced to enable fast accumulations of binary vectors. 
	
	In the case of horizontal summation, CUDA's \texttt{popcount} intrinsic is utilized to count the number of bits set to 1 in a 32-bit integer. As it is known beforehand that only values of 0 or 1 are being accumulated, 8-bit summations can be used during accumulation, with intermediate dumping of the 8-bit accumulators into the output summation vector every seven additions to avoid overflow. 
	
	Figure~\ref{fig:hdextensions} illustrates the bit-packing and horizontal summation HDTorch operations. Vertical summation is accomplished by transposing the input bit-array before performing horizontal summation on the intermediate array. The input array is tiled into subblocks of 128x128 bits, with each subarray assigned to a CUDA warp, which is in charge of transposing the tile. Once all warps have completed their transpositions, the tiles are transposed as they are written back to main memory.
	
	\section{Experimental Setup}
	\label{sec:experimentalsetup}
	To evaluate HDTorch's utility in terms of exploring the HDC design space, we perform a wide range of evaluations in terms of HDC training and inference strategies, datasets, and hyperparameter variations. 
	
	\subsection{Datasets}
	We draw results from experiments performed on five datasets covering a range of sizes, complexities, and use cases. Four datasets are standard HDC benchmarks, while the 5th is a large, highly unbalanced medical dataset, providing a more demanding scenario than the first four datasets.
	
	\subsubsection{Reference HD Computing Benchmarks}
	\label{sec:refDatasets}
	To compare HDTorch performance with HDC implementations available in the literature, we use four benchmark datasets from the online-available UCI repository~\cite{Dua:2019}: 1) ISOLET is an audio dataset containing spoken letters of the English alphabet, 2) MNIST is an image dataset consisting of written digits, 3) UCIHAR is a dataset for classifying human activity from smartphone inertial sensors and, 4) PAMAP is a physical activity dataset containing both inertial sensors and a heart rate monitor. We chose these datasets as they are utilized in previous works demonstrating HD computing on GPUs~\cite{kang_xcelhd_2022, hernandez-cane_onlinehd_2021}, with a wide range of Feature Counts (FC), Class Counts (CC) and Dataset Sizes (DS), as listed in Table~\ref{tab:benchmark_datasets}.
	
	\begin{table}
		\caption{HDTorch benchmark Datasets (FC: feature count, CC: class count, DS: dataset size in number of samples)} 
		\label{tab:benchmark_datasets}
		\renewcommand{\arraystretch}{1.2}%
		\begin{tabular}{p{13mm}p{28mm}p{4mm}p{4mm}p{13mm}}
			\textbf{Dataset} & \textbf{Description} & \textbf{FC} & \textbf{CC} & \textbf{DS $[10^3]$}\\ 
			\hline
			\textbf{PAMAP}    &Activity recognition (IMU + HR) & 31 & 5& $\sim$80\\ 
			\textbf{UCIHAR} & Activity recognition (Smartphone) & 561 &12 & $\sim$10\\ 
			\textbf{ISOLET}  & Voice recognition & 617 & 26  &  $\sim$8\\ 
			\textbf{MNIST} & Handwritten digit recognition & 784 &10 & $\sim$70\\ 
			\textbf{CHB-MIT} & Epilepsy detection & 342 &2 & $\sim$7056\\
		\end{tabular}
	\end{table}
		
	\subsubsection{Epilepsy Benchmark Use-Case}
	\label{subsubsec:CHB-MIT}
	Beyond demonstrating HDTorch's utility on standard HD benchmarks, we wish to demonstrate its ability to enable analysis on large, computationally challenging datasets typical of real-world scenarios such as the ones for continuous monitoring of biomedical data. These datasets contain hundreds of hours of data and are usually highly imbalanced. For this reason, typically only subsets of the entire dataset are utilized for analysis, and it is possible that the results of such studies do not represent the actual performance of the algorithms in the final application (e.g. on a wearable device for long-term detection and monitoring). Thus, we test the ability of HDTorch to explore HD algorithms on the CHB-MIT epilepsy dataset, a widely used open source dataset for epilepsy detection~\cite{shoeb_application_2009, sopic_e-glass_2018, zanetti_robust_2020, pale_multi-centroid_2022}. 
		
	CHB-MIT is an EEG database collected by the Children's Hospital of Boston and MIT. It contains 980 hours of data recorded at 256Hz, consisting of 183 seizures from 24 subjects with medically-resistant seizures ranging in age from 1.5 to 22 years~\cite{shoeb_application_2009, goldberger_ary_l_physiobank_2000}. On average, it has 7.6 $\pm$ 5.8 seizures per subject, and between 23 and 26 channels, of which the 18 channels that are common to all patients are utilized in this work\cmmnt{ (i.e., FP1-F7, F7-T7, T7-P7, P7-O1, FP1-F3, F3-C3, C3-P3, P3-O1, FP2-F4, F4-C4, C4-P4, P4-O2, FP2-F8, F8-T8, T8-P8, P8-O2, FZ-CZ, CZ-PZ)}.
		
	We extract 19 features from each of the 18 channels, similar to~\cite{pale_hyperdimensional_2022}, calculating them on 4 second windows with a moving step of 0.5 seconds. 
	We organize the dataset in two manners before analysis: 1) Subsets of data for each patient that contain all seizure data and 10x more randomly selected non-seizure data (\textit{Fact10}), and 2) the entirety of the dataset divided into approximately one-hour-long segments (\textit{1HSeg}). 
	Previous literature has demonstrated that utilizing different seizure to non-seizure ratios in dataset sub-selections can lead to highly overestimated performance~\cite{pale_multi-centroid_2022}. Thus, in this work, we perform the first (to the best of our knowledge) assessment of HDC performance on the entire database, comparing it to the \textit{Fact10} subset. Evaluation is performed in a time-series split cross-validation (TSCV) approach~\cite{mohammed_time-series_2021}, where only previously acquired data can be used for training, as opposed to, for example, typical leave-one-out cross-validation. More specifically, we perform a cross-validation for each hour-long segment, with segment $n$ as the test set and previous segments 0 to $n-1$ as the training set. This approach also reduces the runtime by approximately half, as it trains on less data in all cross-validations but the last one. 
		
	\subsection{Evaluation Metrics}
	To confirm framework correctness, we evaluate performance in terms of accuracy, memory consumption, and training and inference speedup on the four datasets described in Section~\ref{sec:refDatasets}. 
	We perform these analyses while varying model hyperparameters, specifically, the hypervector dimension $D$ and the online training batch size, or the frequency with which class vectors are updated as a function of the arrival of new training data. 
		
	Concerning the epilepsy dataset use-case, we analyze acceleration due to the utilization of HDTorch for encoding, training, and inference for classical and online HD strategies. With this realized speedup, we are able to evaluate performance of classical and online HDC on the entirety of the CHB-MIT database. Thus, we compare episode detection accuracy for the two previously described dataset selections, \textit{Fact10} and \textit{1HSeg}. Performance is evaluated by concatenating predictions of all cross-validations. We perform moving average smoothing of the predicted labels with a window size of 5s, measure sensitivity (TPR), precision (PPV) and F1 score for both selections, and discuss the differences in performance. 
		
	\subsection{Benchmarking Environment}
		
	Benchmarking and analysis are performed on a server system equipped with a 2-socket, 40-core Intel Xeon Gold 6242R processor capable of frequencies up to 4.1GHz and an NVIDIA Tesla V100 GPU. The software environment consists of Python v3.9.10, PyTorch v1.10.2, and the CUDA driver v11.6. Profiling is accomplished via PyTorch's native profiler, capable of profiling CPU and GPU runtime/memory consumption by function.

	\begin{figure}
		\vspace{1mm}
		\centering
		\includegraphics[width=0.95\linewidth]{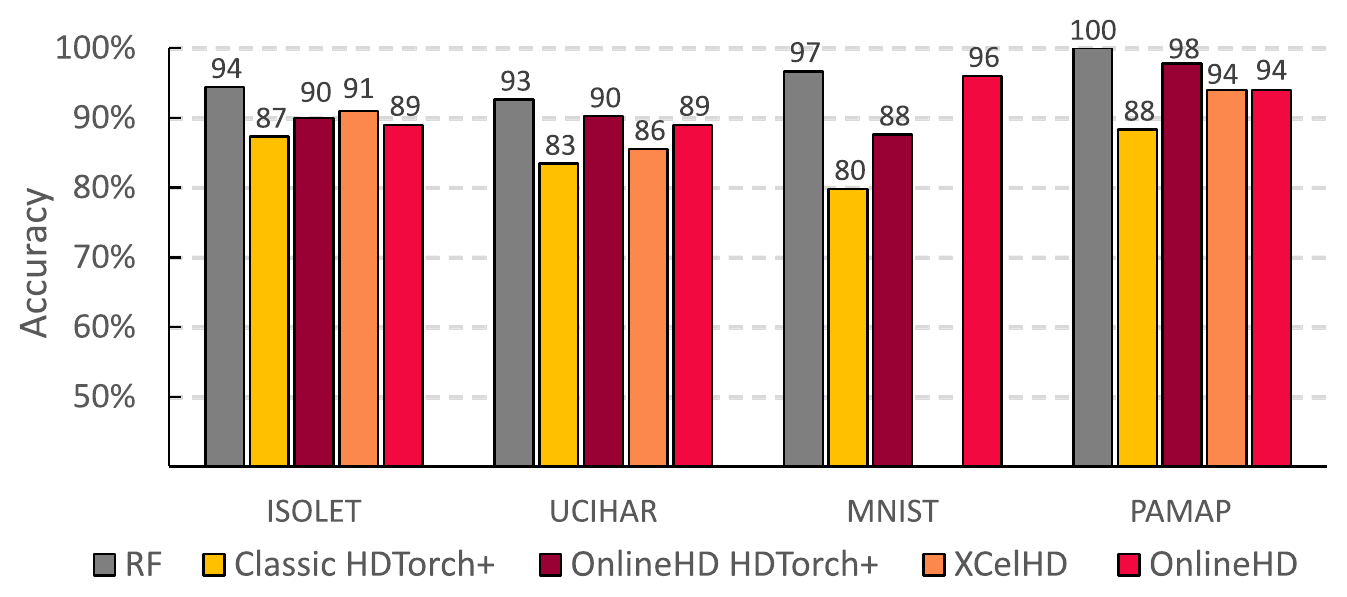}
	    \vspace*{-2mm}
	    \caption{\small{Performances of different HDC implementations, with Random Forest (RF) for reference, on 4 benchmark datasets. }} 
	    \label{fig:AccuracyOtherDatasets}
	\end{figure}
		
	\section{Results}
	\label{sec:results}
	The following sections detail the results of our experiments. For clarity, results utilizing hypervector CUDA extensions are marked as HDTorch+ in the following figures.
	\subsection{Model Accuracy Analysis}
	\label{subsec:accuracy}
		
	Figure~\ref{fig:AccuracyOtherDatasets} shows the accuracy of online and classical HD model implementations with respect to random forest performance. We also compare the online HD approach with two implementations found in the literature. \textit{XCelHD}~\cite{kang_xcelhd_2022} uses a modified TensorFlow implementation of the online HD workflow described in Section~\ref{sec:hdworkflow}, and \textit{OnlineHD}~\cite{hernandez-cane_onlinehd_2021} uses an original HD floating point model with a non-standard encoding approach implemented in PyTorch. 
		
	Online training improves performance in comparison to classical HD for all datasets. Furthermore, our online HD implementation using HDTorch is similar in accuracy to the implementations found the literature. It should be noted that random forest outperforms all HD computing implementations, indicating the necessity for further optimization of HD computing to reach state-of-the-art accuracy results. This situation further motivates the need for efficient design space exploration of HDC algorithms. 
		
		
	\subsection{Time Performance Analysis}
	\label{subsec:timeanalysis}
	\begin{figure}
		\centering
		\includegraphics[trim={1cm 1cm 1cm 1cm},clip,width=\linewidth]{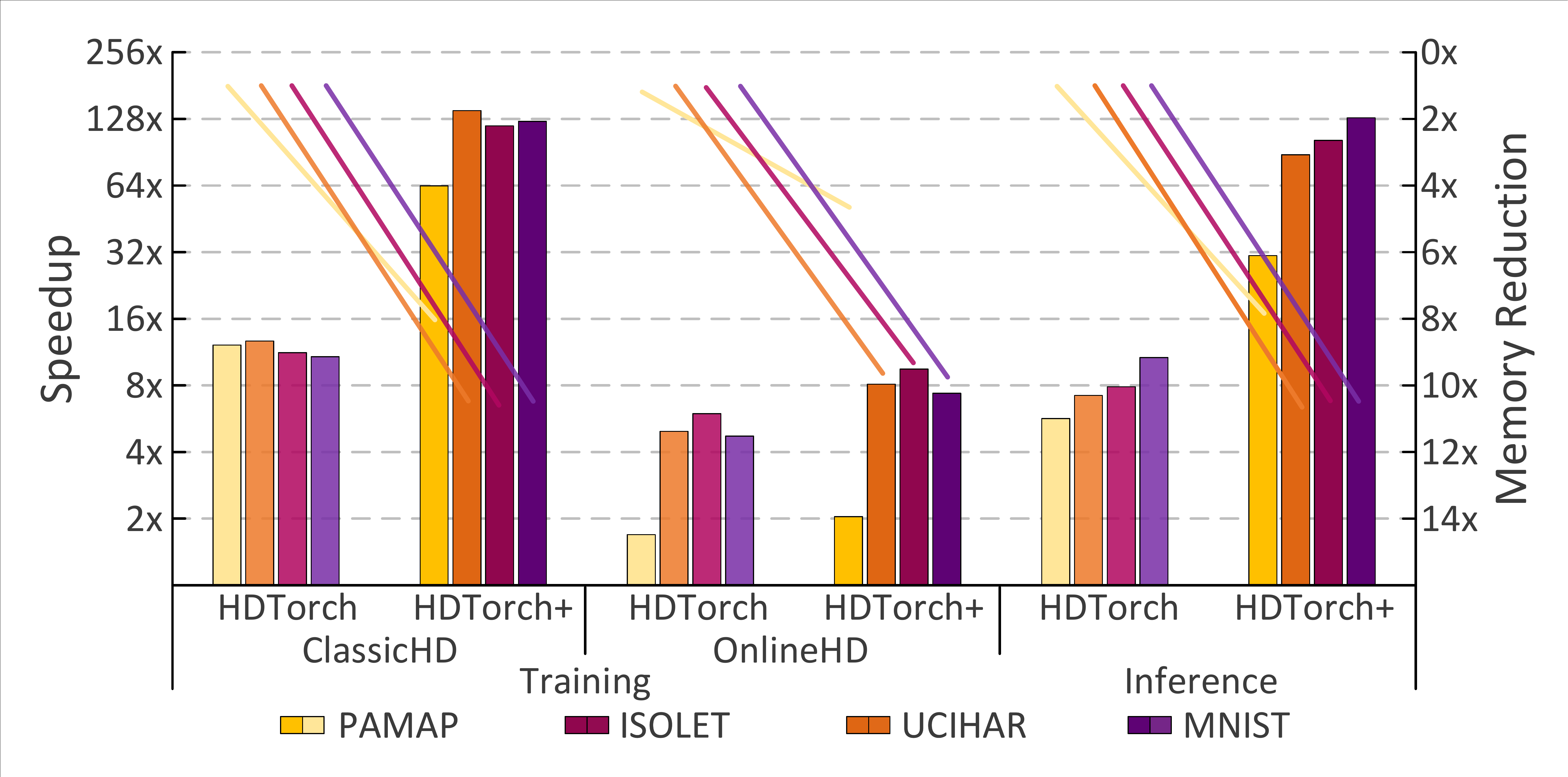}
		\caption{\small{Speedup comparison between different HD learning implementations. 4 different benchmark datasets are used. }} 
		\label{fig:SpeedupOtherDatasets}
	\end{figure}
		
	Figure~\ref{fig:SpeedupOtherDatasets} illustrates runtime accelerations achieved by HDTorch/HDTorch+ in comparison to HDTorch run on the CPU for the four benchmark datasets. Acceleration for training on both classical and online HD, and acceleration for inference (equivalent for both approaches) is illustrated. As can be seen, both HDTorch and HDTorch+ greatly reduce benchmark runtime: HDTorch provides up to a 12.7x/6x/10.7x speedup for classical/online training/inference, respectively, while HDTorch+ improves these gains to 139x/9.5x/130x. It should be noted that online HD speedup is significantly lower as a result of the necessity to re-calculate the class vectors after every datapoint, a challenge addressed in Section~\ref{subsubsec:batch} below. Finally, the observed differences in speedup for different datasets is due to the different dataset feature counts; for example, PAMAP has the smallest number of features and thus the smallest possible speedup due to more limited parallelization opportunities. 
		
	\subsection{Memory Consumption Analysis}
	Figure~\ref{fig:SpeedupOtherDatasets} also illustrates trends in memory consumption across the datasets. Memory values are normalized to the memory usage of running training and inference on the CPU. It can be seen that utilizing HDTorch without HDTorch+ extensions does not reduce memory usage, as the tensors used to store hypervectors are identically sized on either the CPU or GPU. Introducing HDTorch+ extensions, on the other hand, reduces memory consumption by approximately 10x for both training and inference in the aforementioned benchmark datasets. This is due to reducing the memory footprint for hypervectors by 8x by bit packing, and utilizing the more memory efficient HDTorch+ bit-array summation functions, which instantiate fewer intermediate tensors during computation. Once again, PAMAP is an outlier on memory consumption reduction as it has an order of magnitude fewer samples compared to the other datasets.
		
	\begin{figure}

		\centering
		\includegraphics[width=\linewidth,trim={0 0.3cm 0 0},clip]{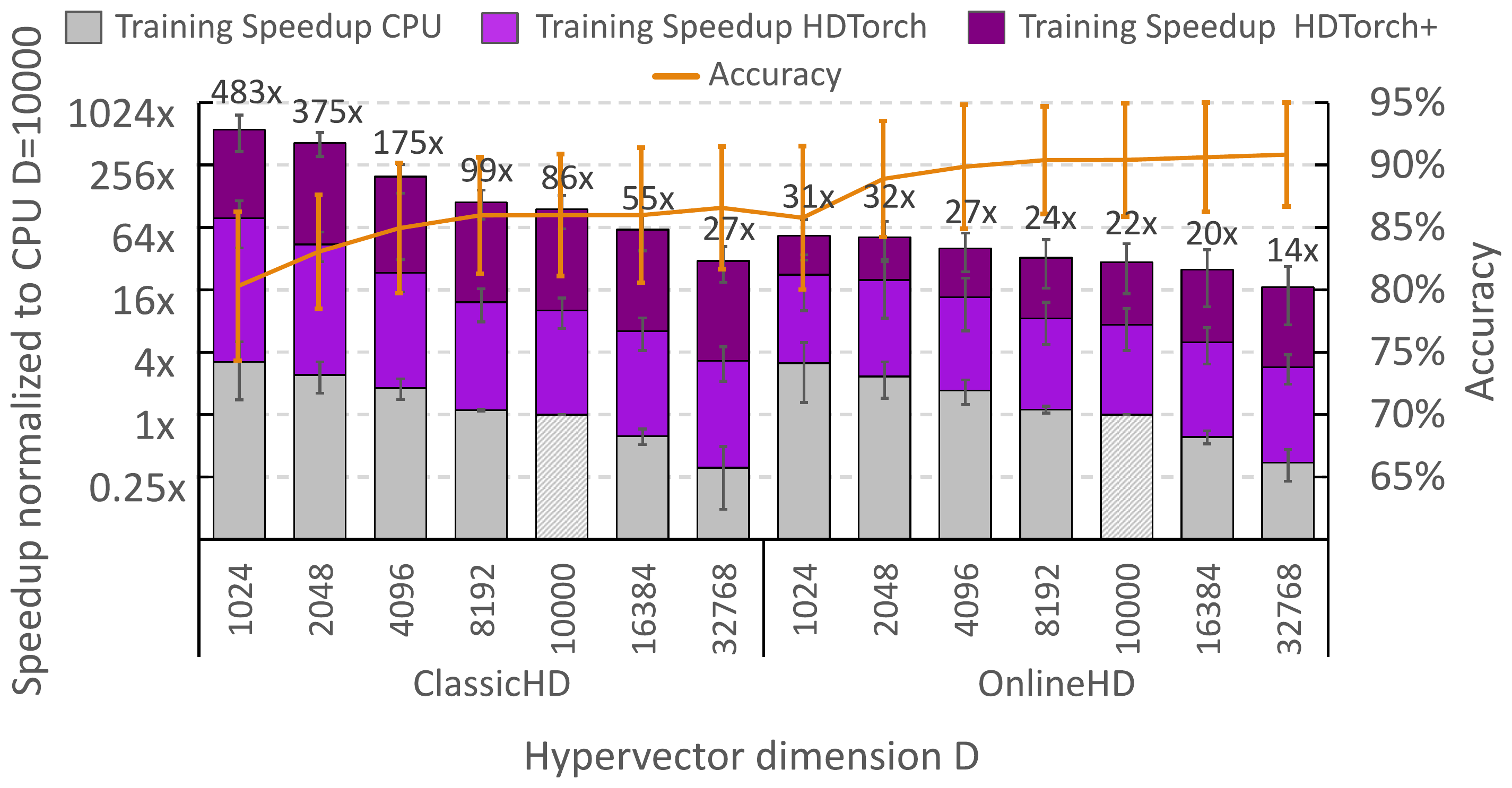}

		\caption{\small{Average speedup for HDTorch-CPU, HDTorch and HDTorch+ for different dimensions of HD vectors on 4 benchmark datasets. Values are normalized to the runtime of the HDTorch model running on CPU with the typical $D$=10000.}} 
		\label{fig:DimensionDCHB-MIT}
	\end{figure}

	\subsection{Parameters Influence}
	We also evaluate the impact of hyperparameter variance on accuracy and runtime for the four benchmarks. We find a trade-off between model complexity/size, accuracy, and runtime. Measured runtime and accuracy values are averaged across the four datasets, and error bars indicate the standard deviation of the averaged values.

	\subsubsection{Hypervector Dimension}

	Figure~\ref{fig:DimensionDCHB-MIT} illustrates the impact on model training time when varying the width $D$ of the class and feature hypervectors between 1024 to 32768 bits. The runtime values are normalized to the runtime of the HDTorch model running on CPU with the typical $D$=10000. As can be seen, reducing $D$ drastically reduces runtime, up to 483x/31x for classical/online learning with HDTorch+, at the cost of an average 6\% decrease in accuracy. On the other hand, increasing $D$ past 10000 provides little improvement to accuracy. However, our results show that even a model with a $D$=32768 runs 27x/14x faster over a CPU model of $D$=10000 for classical/online HD, indicating that if a user would like to test a higher dimensional model on their dataset, HDTorch makes this feasible in a reasonable amount of time.
		
	\subsubsection{Batch Size}
	\label{subsubsec:batch}
	\begin{figure}
		\vspace{1mm}
		\centering
		\includegraphics[width=0.95\linewidth]{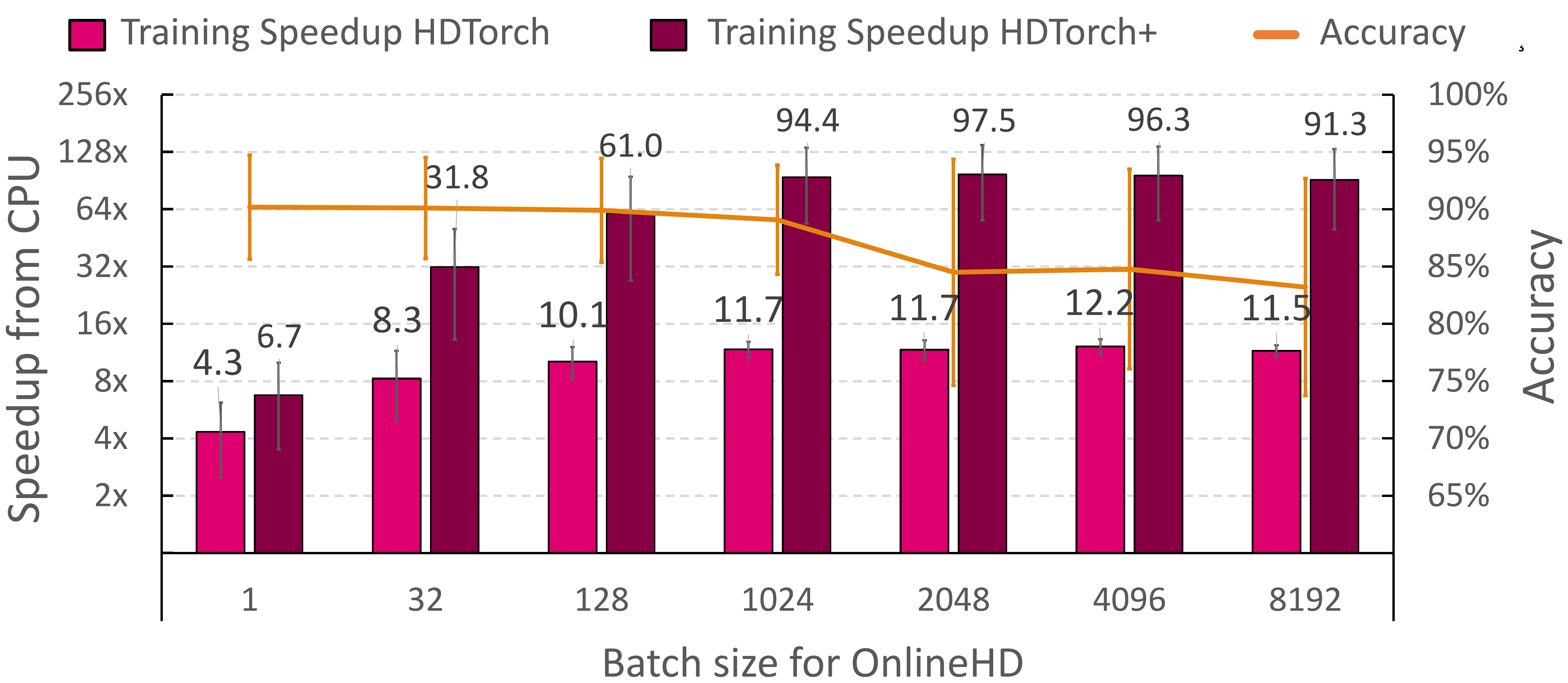}
		\vspace*{-2mm}
		\caption{\small{Acceleration with respect to HDTorch-CPU, compared for HDTorch and HDTorch+ implementations. Comparison is given for training using different batch sizes.}} 
		\label{fig:ParametersBatch}
	\end{figure}
		
	Online HD benefits less from HDTorch acceleration due to the fact that online HD is trained sequentially, with one data sample accumulated into the class vector at a time, and thus cannot be highly parallelized. 
	There is, however, the possibility to improve runtime by only updating class vectors after a batch size of $n$ new datapoints are received. 
	Figure~\ref{fig:ParametersBatch} illustrates batching effects on model runtime and accuracy. It can be seen that batching data significantly decreases training time, especially for HDTorch+, with an average of 68x performance gain achieved across all batch sizes and datasets. This comes at a cost of an average accuracy drop of 7\% at a batch size of 8192. This drop will vary according to dataset complexity; hence, batch size should be tuned to the particular use case of the model.
		
	\subsection{Epilepsy Detection Use Case}
		
	\begin{figure}
		\vspace{1mm}
		\centering
		\includegraphics[width=0.95\linewidth]{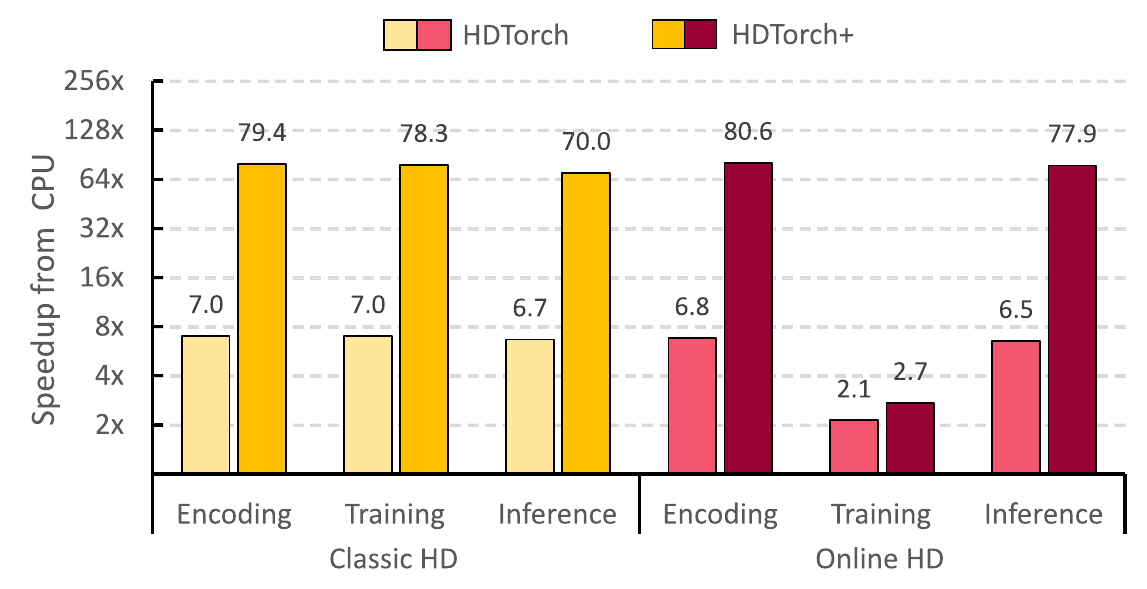}
		\vspace*{-2mm}
		\caption{\small{Acceleration in respect to HDTorch-CPU, compared for HDTorch and HDTorch+ implementations, for encoding stage, training and interference on CHB-MIT dataset.}} 
		\label{fig:SpeedupCHB-MIT}
	\end{figure}
		
	Finally, we test classical HD and online HD implementations on a real-life dataset for epilepsy detection.
	Figure~\ref{fig:SpeedupCHB-MIT} illustrates HDTorch and HDTorch+ speedup with respect to CPU-HDTorch for classical/online HD. Accelerations are shown for the encoding, training, and inference stages.
	For classical HD, HDTorch achieves a 7x speedup for all stages, with HDTorch+ providing an additional 10x gain for each stage, for total speedups of 79x, 78x, and 70x for encoding, training, and inference, respectively. For online HD, performance gains are similar for encoding and inference, whereas for training, gains are significantly reduced, to 2.1x and 2.7x for HDTorch and HDTorch+. This is due to the parallelization constraints described in Section~\ref{subsec:accuracy}.

	\begin{figure}
		\vspace{1mm}
		\centering
		\includegraphics[width=0.95\linewidth]{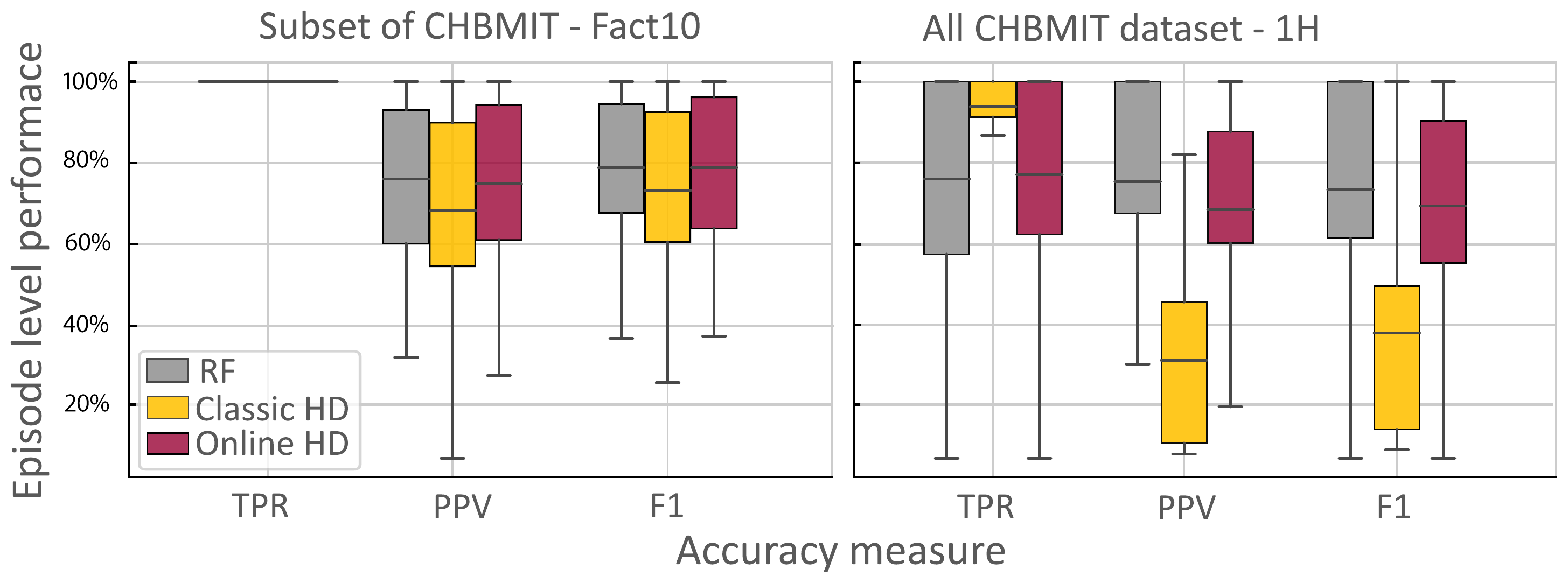}
		\vspace*{-2mm}
		\caption{\small{Epilepsy detection performance comparing training on a subset of data (typical in previous literature) against using the whole CHB-MIT dataset. Boxplots represent performance distribution for all 24 subjects, with mean performance marked as a horizontal line.}} 
		\label{fig:AccuracyCHB-MIT}
	\end{figure}
		
	The speedup achieved by using the HDTorch library enables us to evaluate HDC performance on the entirety of the CHB-MIT database. Evaluation of HDC performance on such a big database enables better insights into performance of HDC algorithms for real-life applications on wearable devices.
	Figure~\ref{fig:AccuracyCHB-MIT} illustrates the detection of epileptic episodes for the two dataset organizations described in Section~\ref{subsubsec:CHB-MIT}, namely, \textit{Fact10} and \textit{1HSeg}.
	For \textit{Fact10}, there is no significant difference between performance of random forest, classical HD and online HD, with all of them detecting almost all seizures with some amount of false positives. On the other hand, training and predicting on the \textit{1HSeg} dataset organization shows a significant decrease in performance for classical HD. While it still detects almost all seizures, it also has significantly more false positives, thus reducing the F1 score. While random forest and online HD miss some seizures, they also avoid such a significant drop in precision and F1 score. Even so, random forest slightly outperforms online HD performance. 
		
	These results indicate two key takeaways. First, analyzing a subset of data may not generalize to performance on the entirety of a dataset, especially when the dataset is large and highly unbalanced, as is the case of the CHB-MIT database. Second, new training strategies are constantly improving HD computing's accuracy, bringing it closer to that of other standard ML models, such as random forest. Both takeaways motivate the necessity for HDTorch, which enables new HDC strategies to be explored on realistic datasets, paving the way for future breakthroughs in the field.
	
	\section{Conclusion}
	\label{sec:conclusion}
	In order to bring HDC performance in line with state-of-the-art ML algorithms and position it as an algorithm for continuous online monitoring for wearables in healthcare, algorithm optimization and design space exploration are necessary. 
	Thus, in this work we have presented HDTorch, the first open-source, PyTorch-based library for HD computing with CUDA extensions for hypervector operations.
		
	On four HDC benchmark datasets, we demonstrated an average 111x/68x training speedup for classical/online HD, respectively, and an average 87x speedup for inference.
	In addition, we have shown that HDTorch's CUDA extensions for HDC operations reduce training/inference memory consumption by up to 10x. HDTorch's utility and flexibility have been demonstrated by analyzing the effects of hypervector dimension and batch size on model accuracy and runtime. 
		
	Finally, the speedup achieved by using the HDTorch library enables us to evaluate HDC performance on the large, highly unbalanced CHB-MIT epilepsy dataset, where we demonstrate that the performance resulting from typical approach of training on a subset of the data does not necessarily generalize to training on the entire dataset. This important observation must be carefully considered when developing future HD models for medical wearable devices.

		

		
	\small
	\printbibliography

@article{kanerva2009hyperdimensional,
  title={Hyperdimensional computing: An introduction to computing in distributed representation with high-dimensional random vectors},
  author={Kanerva, Pentti},
  journal={Cognitive computation},
  year={2009}
}

@article{hernandez-cane_onlinehd_2021,
	title = {OnlineHD: Robust, Efficient, and Single-Pass Online Learning Using Hyperdimensional System},
	journal = {DATE},
	author = {Hernandez-Cane, Alejandro and Matsumoto, Namiko and Ping, Eric and Imani, Mohsen},
	year = {2021},
}

@article{kang_xcelhd_2022,
	title = {XCelHD: An Efficient GPU-Powered Hyperdimensional Computing with Parallelized Training},
	journal = {ASP-DAC},
	author = {Kang, Jaeyoung and Khaleghi, Behnam and Kim, Yeseong and Rosing, Tajana},
	year = {2022},
}

@misc{Dua:2019 ,
    author = "Dua, Dheeru and Graff, Casey",
    year = "2017",
    title = "{UCI} Machine Learning Repository",
    url = "http://archive.ics.uci.edu/ml",
    institution = "University of California, Irvine, School of Information and Computer Sciences"
}

@phdthesis{shoeb_application_2009,
	type = {Thesis},
	title = {Application of machine learning to epileptic seizure onset detection and treatment},
	copyright = {M.I.T. theses are protected by  copyright. They may be viewed from this source for any purpose, but  reproduction or distribution in any format is prohibited without written  permission. See provided URL for inquiries about permission.},
	school = {Massachusetts Institute of Technology},
	author = {Shoeb, Ali Hossam},
	year = {2009},
}

@article{goldberger_ary_l_physiobank_2000,
	title = {PhysioBank PhysioToolkit and PhysioNet},
	journal = {Circulation},
	author = {{Goldberger Ary L.} and {Amaral Luis A. N.} and {Glass Leon} and {Hausdorff Jeffrey M.} and {Ivanov Plamen Ch.} and {Mark Roger G.} and {Mietus Joseph E.} and {Moody George B.} and {Peng Chung-Kang} and {Stanley H. Eugene}},
	year = {2000},
}

@article{pale_multi-centroid_2022,
	title = {Multi-Centroid Hyperdimensional Computing Approach for Epileptic Seizure Detection},
	journal = {Frontiers in Neurology},
	author = {Pale, Una and Teijeiro, Tomas and Atienza, David},
	year = {2022},
}

@article{pale_exploration_2022,
	title = {Exploration of Hyperdimensional Computing Strategies for Enhanced Learning on Epileptic Seizure Detection},
	journal = {arXiv:2201.09759},
	author = {Pale, Una and Teijeiro, Tomas and Atienza, David},
	year = {2022},
	note = {arXiv: 2201.09759},
}

@article{imani_semihd_2019,
	title = {SemiHD: Semi-Supervised Learning Using Hyperdimensional Computing},
	shorttitle = {{SemiHD}},
	journal = {ICCAD},
	author = {Imani, Mohsen and Bosch, Samuel and Javaheripi, Mojan and Rouhani, Bita and Wu, Xinyu and Koushanfar, Farinaz and Rosing, Tajana},
	year = {2019},
}

@article{imani_voicehd_2017,
	title = {VoiceHD: Hyperdimensional Computing for Efficient Speech Recognition},
	doi = {10.1109/ICRC.2017.8123650},
	journal = {ICRC},
	author = {Imani, M. and Kong, D. and Rahimi, A. and Rosing, T.},
}

@article{imani_exploring_2017,
	title = {Exploring Hyperdimensional Associative Memory},
	journal = {HPCA},
	author = {Imani, Mohsen and Rahimi, Abbas and Kong, Deqian and Rosing, Tajana and Rabaey, Jan M.},
	year = {2017},
}

@article{gupta_felix_2018,
	title = {FELIX: Fast and Energy-Efficient Logic in Memory},
	journal = {ICCAD},
	author = {Gupta, Saransh and Imani, Mohsen and Rosing, Tajana},
	year = {2018},
}

@article{imani_hierarchical_2018,
	title = {Hierarchical Hyperdimensional Computing for Energy Efficient Classification},
	journal = {DAC},
	author = {Imani, Mohsen and Huang, Chenyu and Kong, Deqian and Rosing, Tajana},
	year = {2018},
}

@article{wu_brain-inspired_2018,
	title = {Brain-inspired computing exploiting carbon nanotube {FETs} and resistive {RAM}: Hyperdimensional computing case study},
	journal = {ISSCC},
	author = {Wu, Tony F. and Li, Haitong and Huang, Ping-Chen and Rahimi, Abbas and Rabaey, Jan M. and Wong, H.-S. Philip and Shulaker, Max M. and Mitra, Subhasish},
	year = {2018},
}

@article{rahimi_robust_2016,
	title = {A Robust and Energy-Efficient Classifier Using Brain-Inspired Hyperdimensional Computing},
	journal = {ISLPED},
	author = {Rahimi, Abbas and Kanerva, Pentti and Rabaey, Jan M.},
	year = {2016},
}

@article{karunaratne_energy_2021,
	title = {Energy Efficient In-Memory Hyperdimensional Encoding for Spatio-Temporal Signal Processing},
	journal = {TCAS-II: Express Briefs},
	author = {Karunaratne, Geethan and Le Gallo, Manuel and Hersche, Michael and Cherubini, Giovanni and Benini, Luca and Sebastian, Abu and Rahimi, Abbas},
	year = {2021},
}

@article{karunaratne_-memory_2020,
	title = {In-memory hyperdimensional computing},
	journal = {Nature Electronics},
	author = {Karunaratne, Geethan and Le Gallo, Manuel and Cherubini, Giovanni and Benini, Luca and Rahimi, Abbas and Sebastian, Abu},
	year = {2020},
}

@article{imani_revisiting_2021,
	title = {Revisiting HyperDimensional Learning for FPGA and Low-Power Architectures},
	journal = {HPCA},
	author = {Imani, Mohsen and Zou, Zhuowen and Bosch, Samuel and Rao, Sanjay Anantha and Salamat, Sahand and Kumar, Venkatesh and Kim, Yeseong and Rosing, Tajana},
	year = {2021},
}

@article{chang_hyperdimensional_2019,
	title = {Hyperdimensional Computing-based Multimodality Emotion Recognition with Physiological Signals},
	journal = {AICAS},
	author = {Chang, En-Jui and Rahimi, Abbas and Benini, Luca and Wu, An-Yeu Andy},
	year = {2019},
}

@article{rahimi_hyperdimensional_2016,
	author = {Rahimi, A. and Benatti, S. and Kanerva, P. and Benini, L. and Rabaey, J. M.},
	title = {Hyperdimensional biosignal processing: {A} case study for {EMG}-based hand gesture recognition},
	journal = {ICRC},
	year = {2016},
}

@article{burrello_ensemble_2021,
	title = {An Ensemble of Hyperdimensional Classifiers: Hardware-Friendly Short-Latency Seizure Detection With Automatic iEEG Electrode Selection},
	journal = {J-BHI},
	author = {Burrello, Alessio and Benatti, Simone and Schindler, Kaspar and Benini, Luca and Rahimi, Abbas},
	year = {2021},
}

@article{moin_wearable_2021,
	title = {A wearable biosensing system with in-sensor adaptive machine learning for hand gesture recognition},
	journal = {Nature Electronics},
	author = {Moin, Ali and Zhou, Andy and Rahimi, Abbas and Menon, Alisha and Benatti, Simone and Alexandrov, George and Tamakloe, Senam and Ting, Jonathan and Yamamoto, Natasha and Khan, Yasser and Burghardt, Fred and Benini, Luca and Arias, Ana C. and Rabaey, Jan M.},
	year = {2021},
}

@article{benatti_online_2019,
	title = {Online Learning and Classification of EMG-Based Gestures on a Parallel Ultra-Low Power Platform Using Hyperdimensional Computing},
	journal = {TBioCAS},
	author = {Benatti, Simone and Montagna, Fabio and Kartsch, Victor and Rahimi, Abbas and Rossi, Davide and Benini, Luca},
	year = {2019},
}

@article{imani_framework_2019,
	title = {A Framework for Collaborative Learning in Secure High-Dimensional Space},
	journal = {CLOUD},
	author = {Imani, Mohsen and Kim, Yeseong and Riazi, Sadegh and Messerly, John and Liu, Patric and Koushanfar, Farinaz and Rosing, Tajana},
	year = {2019},
}

@article{asgarinejad_detection_2020,
	title = {Detection of Epileptic Seizures from Surface EEG Using Hyperdimensional Computing},
	journal = {EMBC},
	author = {Asgarinejad, Fatemeh and Thomas, Anthony and Rosing, Tajana},
	year = {2020},
}

@article{burrello_laelaps_2019,
	title = {Laelaps: An Energy-Efficient Seizure Detection Algorithm from Long-term Human iEEG Recordings without False Alarms},
	author = {Burrello, Alessio and Cavigelli, L. and Schindler, K. and Benini, L. and Rahimi, Abbas},
	year = {2019},
	journal = {DATE},
}

@article{pale_systematic_2021,
	title = {Systematic Assessment of Hyperdimensional Computing for Epileptic Seizure Detection},
	journal = {EMBC},
	author = {Pale, Una and Teijeiro, T. and Alonso, David Atienza},
	year = {2021},
}

@article{mitrokhin_learning_2019,
	title = {Learning sensorimotor control with neuromorphic sensors: {Toward} hyperdimensional active perception},
	journal = {Science Robotics},
	author = {Mitrokhin, A. and Sutor, P. and Fermüller, C. and Aloimonos, Y.},
	year = {2019}
}

@article{guo_hyperrec_2021,
	title = {HyperRec: Efficient Recommender Systems with Hyperdimensional Computing},
	journal = {ASPDAC},
	author = {Guo, Yunhui and Imani, Mohsen and Kang, Jaeyoung and Salamat, Sahand and Morris, Justin and Aksanli, Baris and Kim, Yeseong and Rosing, Tajana},
	year = {2021},
}

@article{rahimi_hyperdimensional_2020,
	title = {Hyperdimensional Computing for Blind and One-Shot Classification of EEG Error-Related Potentials},
	journal = {Mobile Networks and Applications},
	author = {Rahimi, Abbas and Tchouprina, Artiom and Kanerva, Pentti and Millán, José del R. and Rabaey, Jan M.},
	year = {2020},
}

@article{karunaratne_real-time_2021,
	title = {Real-time Language Recognition using Hyperdimensional Computing on Phase-change Memory Array},
	journal = {AICAS},
	author = {Karunaratne, Geethan and Rahimi, Abbas and Gallo, Manuel Le and Cherubini, Giovanni and Sebastian, Abu},
	year = {2021},
}

@article{yu_artificial_2018,
	title = {Artificial intelligence in healthcare},
	journal = {Nature Biomedical Engineering},
	author = {Yu, Kun-Hsing and Beam, Andrew L. and Kohane, Isaac S.},
	year = {2018},
}

@article{sujith_systematic_2022,
	title = {Systematic review of smart health monitoring using deep learning and Artificial intelligence},
	journal = {Neuroscience Informatics},
	author = {Sujith, A. V. L. N. and Sajja, Guna Sekhar and Mahalakshmi, V. and Nuhmani, Shibili and Prasanalakshmi, B.},
	year = {2022},
}

@article{he2019mlframeworks,
author = {He, Horace},
title = {The State of Machine Learning Frameworks in 2019},
journal = {The Gradient},
year = {2019},
howpublished = {\url{https://thegradient.pub/state-of-ml-frameworks-2019-pytorch-dominates-research-tensorflow-dominates-industry/ } },
}

@article{pale_hyperdimensional_2022,
	title = {Hyperdimensional computing encoding for feature selection on the use case of epileptic seizure detection},
	url = {http://arxiv.org/abs/2205.07654},
	journal = {arXiv:2205.07654},
	institution = {arXiv},
	author = {Pale, Una and Teijeiro, Tomas and Atienza, David},
	year = {2022},
	doi = {10.48550/arXiv.2205.07654},
	note = {arXiv:2205.07654 [cs, eess] type: article},
}

@article{sopic_e-glass_2018,
	title = {e-Glass: A Wearable System for Real-Time Detection of Epileptic Seizures},
	journal = {AISCAS},
	author = {Sopic, Dionisije and Aminifar, Amir and Atienza, David},
	year = {2018},
}

@article{zanetti_robust_2020,
	title = {Robust Epileptic Seizure Detection on Wearable Systems with Reduced False-Alarm Rate},
	journal = {EMBC},
	author = {Zanetti, Renato and Aminifar, A. and Atienza, D.},
	year = {2020},
}

@article{mohammed_time-series_2021,
	title = {Time-Series Cross-Validation Parallel Programming using MPI},
	journal = {ICDABI},
	author = {Mohammed, Abdulrahim and Khedr, Ahmed and AlHaj, Duaa and Al Khalifa, Reem and Alqaddoumi, Abdulla},
	year = {2021},
}
\end{document}